\documentclass[pmlr]{jmlr}

\usepackage{longtable}

\usepackage{booktabs}
\usepackage[load-configurations=version-1]{siunitx} 

\theorembodyfont{\upshape}
\theoremheaderfont{\scshape}
\theorempostheader{:}
\theoremsep{\newline}

\jmlrvolume{XX}
\jmlryear{2019}
\jmlrworkshop{Machine Learning for Health (ML4H) at NeurIPS 2019}

\title[Generative Smoke Removal]{Generative Smoke Removal}

  \author{\Name{Oleksii Sidorov} \Email{oleksiis@fb.com}\\
           \Name{Congcong Wang} \Email{congcong.wang@ntnu.no}\\
            \Name{Faouzi Alaya Cheikh} \Email{faouzi.cheikh@ntnu.no}\\
           \addr The Norwegian Colour and Visual Computing Laboratory\\
                 Norwegian University of Science and Technology
                 }

\editor{Editor's name}

\begin{document}

\maketitle

\begin{abstract}
In minimally invasive surgery, the use of tissue dissection tools causes smoke, which inevitably degrades the image quality. This could reduce the visibility of the operation field for surgeons and introduces errors for the computer vision algorithms used in surgical navigation systems. In this paper, we propose a novel approach for computational smoke removal using supervised image-to-image translation. We demonstrate that straightforward application of existing generative algorithms allows removing smoke but decreases image quality and introduces synthetic noise (grid-structure). Thus, we propose to solve this issue by modification of GAN’s architecture and adding perceptual image quality metric to the loss function. Obtained results demonstrate that proposed method efficiently removes smoke as well as preserves perceptually sufficient image quality.
\end{abstract}

\section{Introduction}
\label{intro}
In laparoscopic surgery, patient's abdomen is visualized by a camera which is inserted into the body through small incisions. High quality of the captured video is necessary to keep a clear visualization for the operating surgeons as well as for navigation systems~\citep{stoyanov2012surgical, andrea2018validation}. However, the perceptual quality can be significantly degraded by smoke caused by such tool as laser ablation. The surgeons' visibility is inevitably impacted by this degradation.    Furthermore, the computer vision technique based surgical navigation systems are mainly designed for clear videos~\citep{andrea2018validation,wang2018liver}, smoke would influence the performance. Therefore, in order to maintain a clear operation field, it becomes necessary to remove the smoke from laparoscopic images by smoke evacuation techniques~\citep{lawrentschuk2010laparoscopic} and by computer vision algorithms~\citep{wang2018variational}. Especially, a real time automatic image 
processing based method is desired which would not introduce any extra hardware to the surgical procedure. Moreover, the algorithm can be embedded to computer assisted surgical navigation workflow easily. 
\par 
The majority of existing smoke removal algorithms are based on simplified physical models or assumptions about input image data which limit their practical application. In this paper, our goal is to avoid using any assumptions or simplified models and perform real-time smoke removal end-to-end. It became possible due to the development of deep learning algorithms oriented on conditional image generation. These algorithms utilize the adversarial process to learn mapping between two image domains in a supervised manner. In case of smoke removal task, these two domains are images with and without smoke correspondingly. 
\par 
Thus, in our work, we perform analysis of image generation errors and propose a new loss function which allows to optimize image quality during the training and produces data without noise and artifacts.  We utilize a set of smoke-free images cast with synthetic smoke to train the network, and further evaluate its performance on real-world data. 
\par 
The remainder of this paper is structured as follows. In Section~\ref{sec:related work}, we review the related work on smoke removal and image generation using GANs. Next, in Section~\ref{sec:method}, we describe our proposed method. Section~\ref{sec:results} presents the training of the network and discusses the experimental results. Finally, the conclusions are drawn in Section~\ref{sec:conclusions}.

\section{Related works}
\label{sec:related work}

\subsection{Laparoscopic Smoke Removal}
In this part, we group the smoke removal methods to traditional approaches and deep learning approaches.  

\paragraph{Traditional approaches:} As dehazing and desmoking problem share some similarity, traditional desmoking approaches~\citep{wang2018variational,kotwal2016joint,baid2017joint,tchakaa2017chromaticity,luo2017vision} follow similar strategy as dehazing approaches. In those literatures, the atmospheric scattering model~\citep{narasimhan2002vision} described in Equation (\ref{eq:scattering}) is widely used. 
\begin{equation}
\label{eq:scattering}
\mathbf{I}(x,y)=\mathbf{J}(x,y)t(x,y)+\mathbf{A}(1-t(x,y)),
\end{equation}
where $\mathbf{I}$ is the observed haze image, $\mathbf{J}$ represents the haze-free image, $\mathbf{A}$ is the global atmospheric light and $t$ is the medium transmission map. In~\citep{kotwal2016joint}, desmoking and denoising problem is formulated to probabilistic graphical model and then it is extended in~\citep{baid2017joint} for desmoking, denoising and specular removal. In~\citep{tchakaa2017chromaticity}, a dark channel prior dehazing method originally proposed in~\citep{he2011single} is modified for smoke removal purpose. In~\citep{luo2017vision}, Luo \textit{et al.} propose to estimate atmospheric veil ($\textbf{A}(1-t(x,y))$) directly instead of calculating $t$. In~\citep{wang2018variational}, Wang \textit{et al.} present a variational method to estimate the atmospheric veil. The approaches proposed in~\citep{luo2017vision,wang2018variational} show promising results, but the performance degraded for dense and heterogeneous smoke. Moreover, all the methods rely on some assumptions, therefore, the methods' performance degenerates when the assumptions are wrong.

\paragraph{Deep Learning approaches:} In~\citep{sabri2018}, Bolkar \textit{et al.} propose the first deep learning desmoking approach. Synthetic dataset created by Perlin noise is generated and used for fine tuning AODNet~\citep{li2017aod}. Later, in~\citep{chen2018unsupervised}, a conditional Generative Adversarial Network is trained by Blender\footnote{ \url{https://www.blender.org/}} generated synthetic dataset for desmoking. These deep learning based methods show promising direction for developing real-time smoke removal algorithms. 
\subsection{Image generation using GANs}
The deconvolutional (``transposed convolutional'') layers of a CNN (Convolutional Neural Network) have made possible the generation of an output of the same size as an input image. However, L1- or L2-loss used as similarity metric leads to a prediction of blurred images. Generative Adversarial Networks (GANs)~\citep{goodfellow2014generative} solve this issue by adding an adversarial loss, implemented as separate CNN with a binary output (discriminator), that allows achieving a photo-realistic quality of a synthesis. Recently, GANs demonstrated state-of-the-art performance in numerous computer vision tasks, such as image mapping~\citep{isola2017image}, video generation~\citep{wang2018video},  segmentation~\citep{luc2016semantic},  inpainting~\citep{iizuka2017globally},
\emph{etc}. Furthermore, a few GAN-based models were proposed recently for image dehazing~\citep{bharath2018single, li2018single}.
\par 
Isola \textit{et al.}~\citep{isola2017image} first demonstrated the great potential of supervised image-to-image translation with the algorithm called pix2pix. It was successfully applied to image colorization, segmentation, generation of images from edges and segmented labels. Pix2pix utilizes L1-loss of U-Net~\citep{ronneberger2015u} shaped generator together with an adversarial loss of discriminator. On the other hand, Perceptual Adversarial Network (PAN)~\citep{wang2018perceptual} and pix2pixHD~\citep{wang2017high}, apply similar technique complemented by “perceptual loss”. The latter was first presented by Johnson~\textit{et al.}~\citep{johnson2016perceptual} with an objective to cover not only a pixel-wise similarity but also a similarity of high-level features. A perceptual loss is created as a concatenation of activations extracted from different layers of pretrained VGG-16 (pix2pixHD) or directly from discriminator (PAN). Nevertheless, despite being called “perceptual”, this metric has no relation to the human visual system that we discuss further in the text. 
\par 
While pix2pix and PAN learn mapping in a supervised manner, algorithms like CycleGAN~\citep{zhu2017unpaired}, DiscoGAN~\citep{kim2017learning}, and UNIT~\citep{liu2017unsupervised} utilize cycle loss which allows using unpaired data. This approach is not covered in the scope of this work, but it has high potential to be applied for smoke removal when trained using unrelated batches of images with and without smoke.

\section{Methodology}
\label{sec:method}

\begin{figure}[b!]
\floatconts
{fig:artifacts}
{\caption{(a) input image; (b) raw PAN output; (c) a magnitude spectrum of the Fourier transform of (b); (d) output generated by PAN with added MS-SSIM loss.}}
{\subfigure[]{\includegraphics[width=0.45\linewidth]{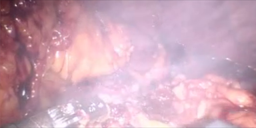}} \quad
\subfigure[]{\includegraphics[width=0.45\linewidth]{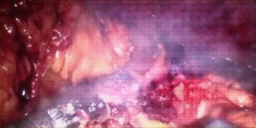}} 
\subfigure[]{\includegraphics[width=0.45\linewidth]{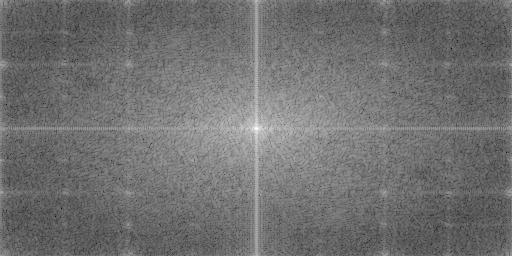}} \quad
\subfigure[]{\includegraphics[width=0.45\linewidth]{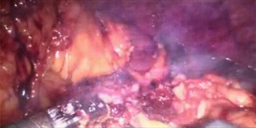}}
}
\end{figure}

\subsection{Model architecture and loss function}
\label{subsec:metod}
As a baseline for the implementation of our approach, we used PAN proposed by Wang \textit{et al.}~\citep{wang2018perceptual}. It has similar architecture to pix2pix except it employs a more efficient loss function. Fig.~\ref{fig:artifacts} (b) illustrates the main drawback of the results obtained when PAN is applied to desmoking straightforward: strongly manifested periodic noise (grid structure). It also manifests in Fourier spectrum of the images (Fig.~\ref{fig:artifacts} (c)). Considering that standard approach to removing spatial periodic noise is to apply image filtering in the frequency space, we designed a mask which maximally covers undesired peaks. Nevertheless, this approach did not improve the image quality significantly, even with various shapes of the mask tested.

The failure of classical image processing in Fourier domain and the desire to keep the model end-to-end motivated us to improve the original PAN algorithm. Since the main issue is the quality of the output images (structural artifacts), the logical step was to add perceptual image quality metric to the loss function and minimize it during training. Peak Signal-to-Noise Ratio (PSNR) and Structural Similarity Index (SSIM)~\citep{wang2004image} are two most commonly used full-reference image quality metrics. The significant advantage of them compared to less popular image quality metrics is their differentiability which allows to use them for gradient computation. However, PSNR (Eq. (\ref{eq:psnr})) was proven~\citep{zhang2012comprehensive} to not correlate well with human perception. Moreover, it is highly related to L2 metric (RMSE) which is not suitable for image translation since it produces blurry results~\citep{isola2017image}.
\begin{equation}
\label{eq:psnr}
    PSNR=20log_{10}(max(I))-10log_{10}MSE
\end{equation}
 SSIM is a perceptual image similarity metric which was proposed as an alternative to MSE (Mean Square Error) and PSNR in order to increase correlation with subjective evaluation. For original and reconstructed images $I$ and $J$, SSIM is defined as:
\begin{equation}
    SSIM(I,J)=\frac{(2\mu _{I}\mu_{J}+c_{1})(2\sigma _{IJ}+c_{2})}{(\mu_{I}^{2}+\mu_{J}^{2}+c_{1})(\sigma_{I}^{2}+\sigma_{J}^{2}+c_{1})},
\end{equation}
where $\mu_I$, $\mu_J$ and $\sigma_I$, $\sigma_J$ are mean and variance of images $I$ and $J$ correspondingly, while $\sigma _{IJ}$ is covariance of the images. Therefore, corresponding loss function was defined as:
\begin{equation}
    L_{SSIM}=-mean(SSIM(I,J)).
\end{equation}
Since SSIM changes in range 0 $\sim$ 1, where higher value corresponds to higher similarity, it has to be inverted in order to join it with other losses which are minimized during an optimization. 

\begin{figure}[t!]
\floatconts
{fig:model}
{\caption{PAN Framework complemented by MS-SSIM image quality loss. ``nAsB" denotes A filters of stride B. The detailed description of the layers is in the text.}}
{\includegraphics[width=1\linewidth]{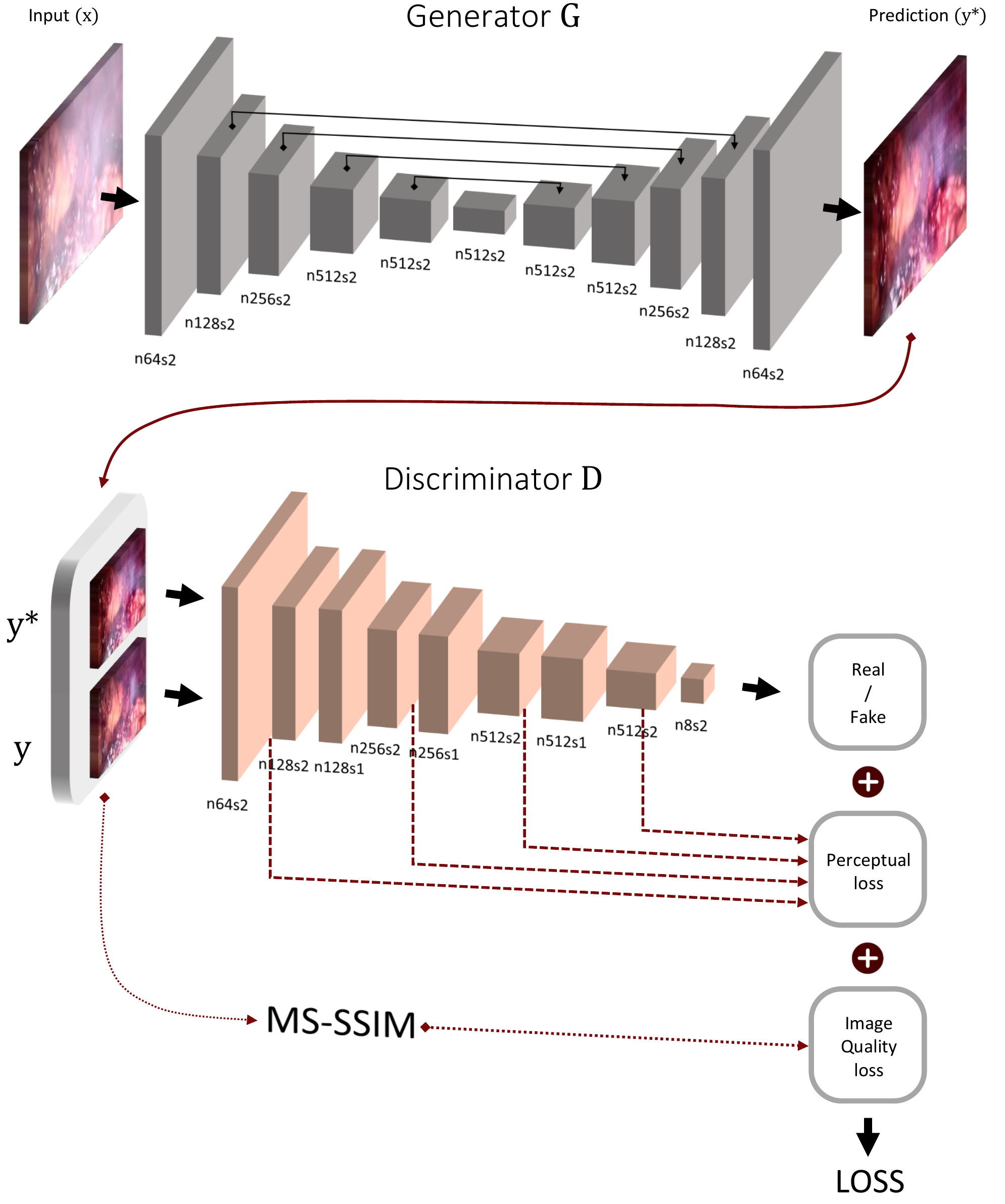}}
\end{figure}

While SSIM processes windows of specified size $n\times n$, its extension Multi-Scale SSIM (MS-SSIM)~\citep{zhang2012comprehensive} takes into account windows of different sizes and allows covering a bigger range of spatial frequencies that lead to better results. So, in addition to SSIM we also used MS-SSIM loss which is computed in the analogous way. 

The complete framework of the algorithm is illustrated in Figure~\ref{fig:model}. The generator $G$ is a U-Net-like convolutional network with skip connections. The discriminator $D$ is a conventional CNN-based binary classifier. The encoding layers of both $G$ and $D$ consist of $Convolution$ layer followed by $BatchNorm$, and $LeakyReLU$. The size of filters in $Convolution$ layer is $3\times3$, whereas their number and stride are marked in Figure~\ref{fig:model}. The decoding layers follows the same template but with $DeConvolution$ of size $4\times4$ instead of $Convolution$. The complex loss function is constructed as a linear combination of the SSIM (MS-SSIM) loss, “perceptual” loss extracted from the discriminator's layers, and an adversarial loss.

\subsection{Implementation Details}
The model is implemented in Pytorch 0.4 and its code is publicly accessible\footnote{ \url{https://github.com/acecreamu/ssim-pan}}. The training was performed using single 4GB GTX980 GPU. Original images were re-sized and zero-padded in order to match 256x256 input size. In our experiments we used ADAM optimizer with a learning rate of 0.0002 and momentum 0.5, batch size 4, and 50 training epochs. All the other parameters' values can be found in the code provided.

\section{Results and discussion}
\label{sec:results}

\subsection{Dataset}
There exist no labeled datasets for desmoking. Therefore, the synthetic dataset from~\citep{wang2019multiscale}~\footnote{ \url{http://hamlyn.doc.ic.ac.uk/vision/}} is used to train our network. Manually selected smoke-free images from~\citep{chen2018unsupervised} are used as the groundtruth images, then Adobe Photoshop\footnote{ \url{https://www.adobe.com/products/photoshop.html}} is used to render clouds to simulate the smoke images appearances. As a result, 7,500 smoke-free images with smoke of three different densities produced 22,500 synthetic image pairs which were used for training. The evaluation, however, was performed using both types of images: with synthetic smoke (ground truth available) and with real smoke (no ground truth)\footnote{ The data with real smoke has been captured dynamically during surgery.}.

\begin{table}[b!]
\floatconts
{table:1}
{\caption{Quantitative evaluation results.}}
{
 \begin{tabular}{l p{1cm} c c p{1cm} c c} 
 \hline\hline
&  \multicolumn{1}{c}{} & \multicolumn{2}{c}{CIEDE2000} & \multicolumn{1}{c}{} & \multicolumn{2}{c}{RMSE} \\ 
& & mean & std & & mean & std \\ 
 \hline
 RDCP~\citep{tchakaa2017chromaticity} & & 12.9 & 1.32 & & 35.0 & \textbf{4.55} \\ 
 DCP~\citep{he2011single}  & & 11.5 & 2.13 & & 40.6 & 6.98 \\
 VAR~\citep{wang2018variational}  & & 10.8 & 2.18 & & 38.1 & 8.00 \\
 EVID~\citep{galdran2015enhanced} & & 7.41 & 1.40 & & 24.2 & 4.81 \\
 Proposed & & \textbf{3.89} & \textbf{1.15} & &\textbf{4.6} & 4.71 \\ 
 \hline\hline
 \end{tabular}
 }
\end{table}

\subsection{Experimental results}
According to the availability of the source code and the suitability of it for smoke removal, we compare the proposed method with four following  methods: physical model based dark channel prior (DCP)~\citep{he2011single} and refined dark channel prior (R-DCP)~\citep{tchakaa2017chromaticity} methods, mild physical constraint based variational approach EVID~\citep{galdran2015enhanced} and recently proposed desmoking approach VAR~\citep{wang2018variational}. 

300 real images with added synthetic smoke have been used for quantitative evaluation due to the direct availability of ground truth. As shown in Table~\ref{table:1}, we report results in terms of colorimetric difference CIEDE2000~\citep{doi:10.1002/col.1049} (which describes accuracy of color reconstruction for human visual system) as well as RMSE of pixel values (which is important for computational algorithms). Our proposed method outperforms the other approaches in terms of CIEDE2000 and RMSE.

\begin{figure*}[t!]
\floatconts
{fig:result_pan_qualitative}
{\caption{Qualitative comparison. (a) Input smoke images and desmoked ones by: (b) DCP~\citep{he2011single}, (c)  VAR~\citep{wang2018variational}, (d) EVID~\citep{galdran2015enhanced}, (e) proposed method. Zoom is required.}}
{\includegraphics[width=1\linewidth]{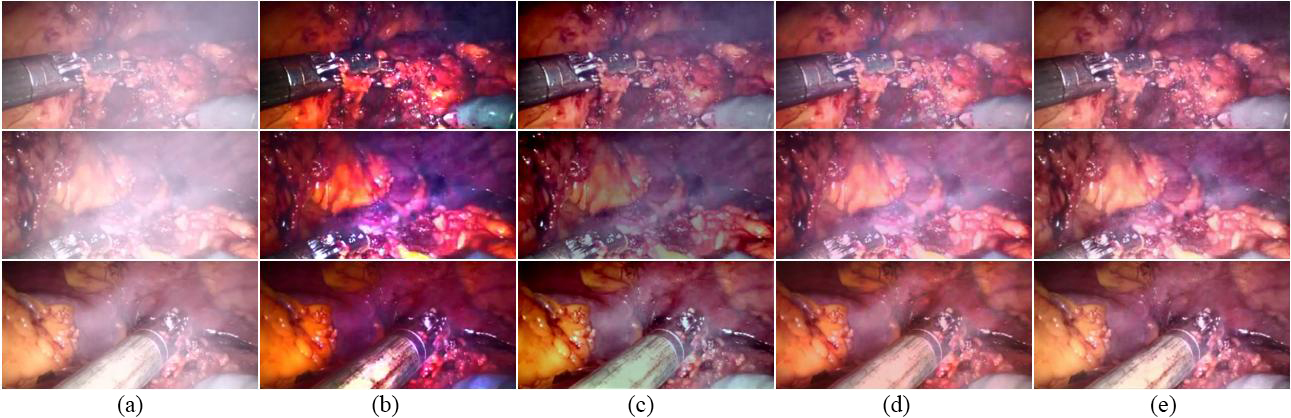}}
\end{figure*}

\begin{figure}[hb!]
\floatconts
{fig:votes}
{\caption{The results of perceptual evaluation by real surgeons.}}
{\includegraphics[width=1.05\linewidth]{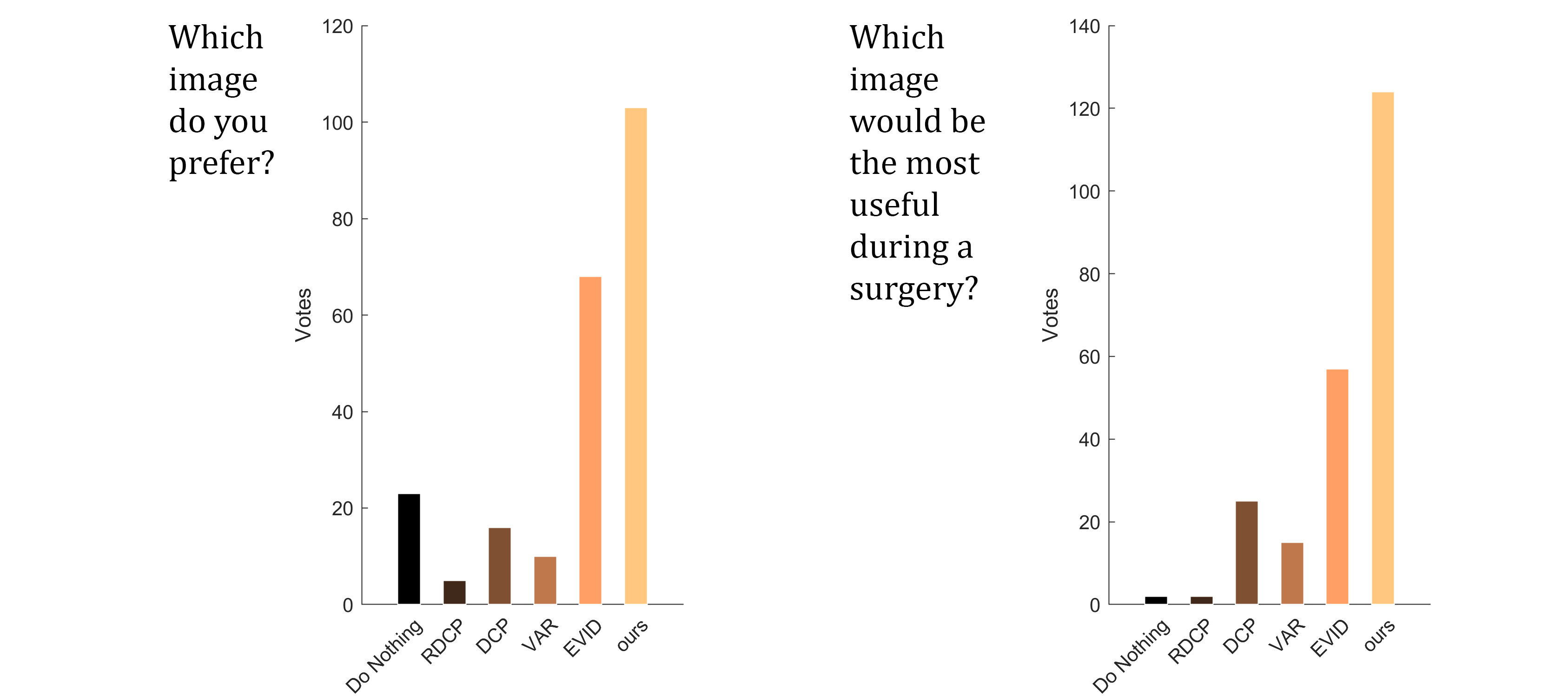}}
\end{figure}

Qualitative result obtained from real smoke images is illustrated in Fig.~\ref{fig:result_pan_qualitative}. It can be clearly seen that the proposed approach outperforms all the other techniques and produces significant visibility enhancement even in the cases with very dense smoke. Another beneficial property is preserving of color information similar to original. Additional improvement of the results is expected in case of using larger and more diverse dataset for training. 

\paragraph{Perceptual evaluation.} The real data does not contain ground truth smoke-free images which makes quantitative comparison of the results troublesome. However, since our primary goals is to enhance visual data used by clinicians for simplification of the surgery process, the most relevant evaluation possible is a perceptual experiment with real doctors. This motivated us to gather subjective responses and define which algorithm is more likely to be chosen for real application.

Surgeons' time is expensive, so we designed our experiment as a short online-survey for the sake of easier accessibility and reaching a larger number of participants. The survey is available online\footnote{ \url{https://www.surveymonkey.com/r/2XL9JPH}} and can be used to find more outputs and evaluate them personally. The participant base consisted of 45 surgeons who kindly followed email-invitation. Each trial presented 10 image-choice questions split into two questions: \emph{``Which image do you prefer?"} and \emph{``Which image would be the most useful during a surgery?"}.  Figure~\ref{fig:votes} illustrates that our approach has earned the largest number of votes in both tasks, even though  EVID method by \citep{galdran2015enhanced} is a strong competitor.

\begin{figure}[t!]
\floatconts
{fig:result_other}
{\caption{The comparison of original and modified versions of PAN illustrated on other datasets.}}
{\includegraphics[width=1\linewidth]{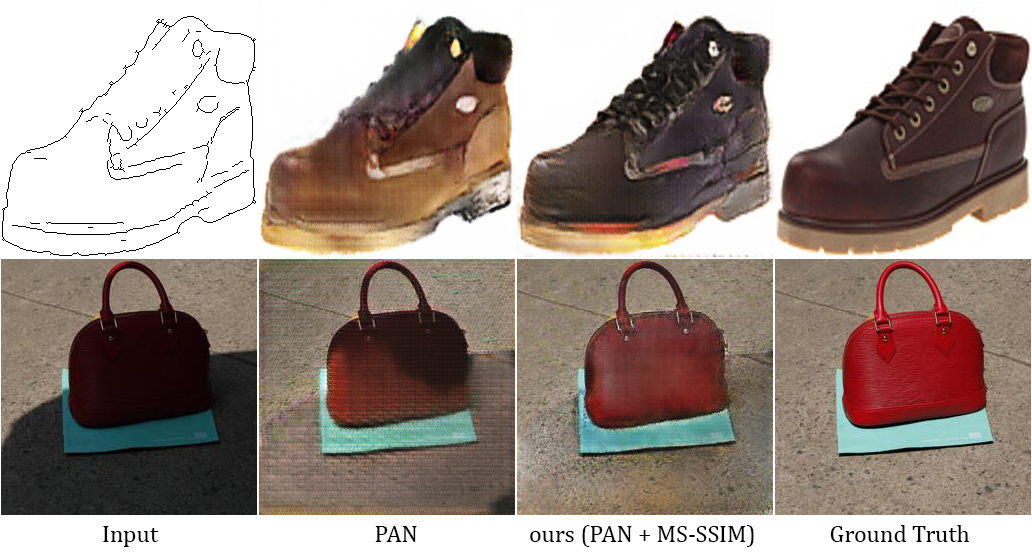}}
\end{figure}

\subsection{Discussion} 
In this paper, we show how image quality metrics can improve the results of supervised image-to-image translation in medical domain. This is not a unique case. The usage of image quality metrics in deep learning was first discussed by Dosovitskiy \textit{et al.}~\citep{dosovitskiy2016generating}. Following works~\citep{snell2017learning,zhao2017loss} illustrated its application to image restoration and super-resolution. However, it has never been applied to GANs where it is especially useful. From the other side, Odena \textit{et al.}~\citep{odena2016deconvolution} relate above-mentioned artifacts (discussed in~\ref{subsec:metod}) to the uneven overlap of deconvolutional filters and propose to solve it by preliminary resizing of an image. Isola\footnote{https://github.com/junyanz/pytorch-CycleGAN-and-pix2pix/issues/78\#issuecomment-322908732} stated that this operation applied to pix2pix can increase training time up to 4 times. On the other hand, our solution increases time of one training epoch only by 13\%, which is annihilated by faster convergence in a fewer number of epochs. 

The additional results of applying the proposed method to the datasets from other domains (edge2shoes~\citep{isola2017image} and SRD~\citep{qu2017deshadownet}) are demonstrated in Fig.~\ref{fig:result_other} and compared to the output of the original PAN from the same epoch. As can be seen, the proposed approach achieves significantly better perceptual image quality regardless of the domain of application.

\section{Conclusions}
\label{sec:conclusions}
In this work, we present the novel GAN-based approach for smoke removal from laparoscopic images. We show how the end-to-end model trained on synthetic data can demonstrate a remarkable performance on real-world images. We used the standard pix2pix-like architecture complemented by ``perceptual" loss and MS-SSIM loss to obtain an effective image enhancement method which is useful for clinicians operating surgery as well as for computer vision algorithms.

Further development may include gathering larger real-world dataset, modification of the code for a real-time application, and application of similar unsupervised methods (CycleGAN, UNIT, \emph{etc.}).


\bibliography{refs}






\end{document}